\documentclass[conference]{IEEEtran}
\IEEEoverridecommandlockouts
\usepackage{cite}
\usepackage{amsmath,amssymb,amsfonts}
\usepackage{algorithmic}
\usepackage{graphicx}
\usepackage{textcomp}
\usepackage{xcolor}

\usepackage{booktabs} 
\usepackage{multirow} 
\usepackage{siunitx}
\usepackage[hidelinks]{hyperref}
\usepackage{listings}
\usepackage{orcidlink}

\def\BibTeX{{\rm B\kern-.05em{\sc i\kern-.025em b}\kern-.08em
    T\kern-.1667em\lower.7ex\hbox{E}\kern-.125emX}}
\begin{document}

\title{BLUEmed: Retrieval-Augmented Multi-Agent Debate for Clinical Error Detection}

% \author{
% \IEEEauthorblockN{
% Saukun Thika You\IEEEauthorrefmark{1},\orcidlink{0009-0009-6437-9017},
% Nguyen Anh Khoa Tran\IEEEauthorrefmark{1},\orcidlink{0009-0005-8687-0668},
% Wesley K. Marizane\IEEEauthorrefmark{1},\orcidlink{0009-0007-6214-0286},\\
% Hanshu Rao\IEEEauthorrefmark{1}\orcidlink{0009-0009-4782-4748},
% Qiunan Zhang\IEEEauthorrefmark{2}\orcidlink{0009-0009-0606-8632},
% Xiaolei Huang\IEEEauthorrefmark{1}\orcidlink{0000-0003-0478-8715}
% }

% \IEEEauthorblockA{
% \IEEEauthorrefmark{1}Department of Computer Science,\\
% University of Memphis, Memphis, United States\\
% \texttt{\{syou,ntran5,wkmrzane,hrao,xiaolei.huang\}@memphis.edu}
% }
% \IEEEauthorblockA{
% \IEEEauthorrefmark{2}Department of Computer Information Systems,\\
% Alabama State University, Montgomery, United States\\
% \texttt{qzhang@alasu.edu}
% }

% \IEEEauthorblockN{1\textsuperscript{st} Given Name Surname}
% \IEEEauthorblockA{\textit{dept. name of organization (of Aff.)} \\
% \textit{name of organization (of Aff.)}\\
% City, Country \\
% email address or ORCID}
% \and
% \IEEEauthorblockN{2\textsuperscript{nd} Given Name Surname}
% \IEEEauthorblockA{\textit{dept. name of organization (of Aff.)} \\
% \textit{name of organization (of Aff.)}\\
% City, Country \\
% email address or ORCID}
% \and
% \IEEEauthorblockN{3\textsuperscript{rd} Given Name Surname}
% \IEEEauthorblockA{\textit{dept. name of organization (of Aff.)} \\
% \textit{name of organization (of Aff.)}\\
% City, Country \\
% email address or ORCID}
% }

\author{\IEEEauthorblockN{Saukun Thika You}
\IEEEauthorblockA{\textit{Department of Computer Science} \\
\textit{University of Memphis}\\
Memphis, United States\\
syou@memphis.edu \orcidlink{0009-0009-6437-9017}}
\and
\IEEEauthorblockN{Nguyen Anh Khoa Tran}
\IEEEauthorblockA{\textit{Department of Computer Science} \\
\textit{University of Memphis}\\
Memphis, United States\\
ntran5@memphis.edu \orcidlink{0009-0005-8687-0668}}
\and
\IEEEauthorblockN{Wesley K. Marizane}
\IEEEauthorblockA{\textit{Department of Computer Science} \\
\textit{University of Memphis}\\
Memphis, United States\\
wkmrzane@memphis.edu \orcidlink{0009-0007-6214-0286}}
\and
\IEEEauthorblockN{Hanshu Rao}
\IEEEauthorblockA{\textit{Department of Computer Science} \\
\textit{University of Memphis}\\
Memphis, United States\\
hrao@memphis.edu \orcidlink{0009-0009-4782-4748}}
\and
\IEEEauthorblockN{Qiunan Zhang}
\IEEEauthorblockA{\textit{Department of Computer Information Systems} \\
\textit{Alabama State University}\\
Montgomery, United States \\
qzhang@alasu.edu \orcidlink{0009-0009-0606-8632}}
\and
\IEEEauthorblockN{Xiaolei Huang}
\IEEEauthorblockA{\textit{Department of Computer Science} \\
\textit{University of Memphis}\\
Memphis, United States\\
xiaolei.huang@memphis.edu \orcidlink{0000-0003-0478-8715}}
}

\maketitle

\begin{abstract}
Terminology substitution errors in clinical notes, where one medical term is replaced by a linguistically valid but clinically different term, pose a persistent challenge for automated error detection in healthcare.
We introduce BLUEmed,\footnote{Our implementation is available at \url{https://github.com/Khoa-BOB/BLUEmed}.} a multi-agent debate framework augmented with hybrid Retrieval-Augmented Generation (RAG) that combines evidence-grounded reasoning with multi-perspective verification for clinical error detection.
BLUEmed decomposes each clinical note into focused sub-queries, retrieves source-partitioned evidence through dense, sparse, and online retrieval, and assigns two domain expert agents distinct knowledge bases to produce independent analyses; when the experts disagree, a structured counter-argumentation round and cross-source adjudication resolve the conflict, followed by a cascading safety layer that filters common false-positive patterns.
We evaluate BLUEmed on a clinical terminology substitution detection benchmark under both zero-shot and few-shot prompting with multiple backbone models spanning proprietary and open-source families.
Experimental results show that BLUEmed achieves the best accuracy (69.13\%), ROC--AUC (74.45\%), and PR--AUC (72.44\%) under few-shot prompting, outperforming both single-agent RAG and debate-only baselines.
Further analyses across six backbone models and two prompting strategies confirm that retrieval augmentation and structured debate are complementary, and that the framework benefits most from models with sufficient instruction-following and clinical language understanding.
% This and the IEEEtran.cls file define the components of your paper [title, text, heads, etc.]. 
% *CRITICAL: Do Not Use Symbols, Special Characters, Footnotes, or Math in Paper Title or Abstract.
\end{abstract}

% keywords

% The conference will accept both regular and short papers. 
% Regular papers (10 pages including references) will describe mature ideas, where a substantial amount of implementation, experimentation, or data collection and analysis has been completed. 
% Short papers (6 pages including references) will describe innovative ideas, where preliminary implementation and validation work have been conducted.
% NOTE: Supplementary materials are not permitted. All content necessary for review must be included within the page limits of the paper.

\section{Introduction}
\label{sec:intro}

% ===== P1 =====
Clinical notes record diagnoses, treatment plans, and patient histories in Electronic Health Records (EHR), and their accuracy directly affects clinical decision-making and patient safety~\cite{alotaibi2017impact,rao2026scoping}.
A recurring source of documentation error is \textit{terminology substitution}, in which one medical term is replaced by another that is linguistically valid but clinically different.
For example, substituting ``atrial fibrillation'' with ``atrial flutter,'' or ``metformin'' with ``methotrexate,'' can lead to misdiagnosis or adverse drug events~\cite{bates2007preventing}.
Unlike typographical or grammatical errors, terminology substitutions preserve syntactic correctness and often appear contextually coherent.
Standard spell-checkers and grammar tools therefore cannot detect them.
Identifying such errors requires domain knowledge and the ability to verify clinical assertions against medical evidence, which makes automated detection difficult.

% ===== P2 =====
Large language models (LLMs) have enabled new approaches to clinical text analysis, and recent work on medical error detection has followed two main paradigms.
The first is \textit{retrieval-augmented generation} (RAG)~\cite{lewis2020retrieval, gao2023retrieval}, where a single LLM retrieves relevant passages from a medical knowledge base and produces a classification grounded in the retrieved evidence.
Anchoring outputs in external sources reduces LLM hallucination~\cite{shuster2021retrieval}.
The second paradigm is \textit{multi-agent debate}~\cite{du2023improving, liang2024encouraging}, where multiple LLM agents independently analyze the input, exchange counterarguments, and reach a joint decision through structured argumentation.
This form of interaction has been shown to improve factual accuracy and reasoning consistency~\cite{chan2023chateval, xiong2023examining}.
Earlier methods based on rule-based pattern matching~\cite{uzuner2009recognizing} and supervised classification~\cite{shivade2014review} have also been applied to clinical error detection, but they are limited in handling semantic-level substitutions.

% ===== P3 =====
Both paradigms have clear limitations when applied to terminology substitution detection in isolation.
Single-agent RAG relies on one retrieval pass and one reasoning path, making it sensitive to retrieval noise and bias in the knowledge source~\cite{fang2024enhancing,yoran2024making,yu2024chain}.
When the retrieval pool contains ambiguous or incomplete information, the model has no mechanism to challenge its own interpretation~\cite{asai2024self,niu2024ragtruth}.
Because both the retrieved evidence and the reasoning trajectory originate from the same source, any bias in the knowledge base propagates directly into the final prediction without an independent check~\cite{hwang2025retrieval}.
Empirical results further show that single-agent models are sensitive to in-context exemplar selection: few-shot demonstrations can cause overly conservative predictions that reduce recall~\cite{zhao2021calibrate,ru2024ragchecker}.
Multi-agent debate without retrieval augmentation has the opposite problem. Without external evidence to constrain reasoning, debate agents tend to over-flag errors~\cite{kenton2024scalable}. 
They achieve high recall by classifying most notes as incorrect, but at the cost of low precision~\cite{cormack2009reciprocal}.
In summary, retrieval augmentation provides factual grounding but lacks multi-perspective verification, while multi-agent debate enables adversarial reasoning but lacks evidential constraints.
No existing framework unifies these two capabilities or includes a mechanism to reduce the false-positive patterns that arise in multi-agent clinical reasoning.

% ===== P4 =====
To address these limitations, we propose \textbf{BLUEmed}, a multi-agent debate framework augmented with hybrid Retrieval-Augmented Generation for detecting terminology substitution errors in clinical notes (Figure~\ref{Method_Framework_Figure}).
BLUEmed separates evidence retrieval, clinical reasoning, and adjudication into three modules.
First, a \textit{Hybrid RAG Module} decomposes each clinical note into focused sub-queries and combines dense, sparse, and online retrieval through Reciprocal Rank.
The retrieved evidence is partitioned by source: two \textit{Domain Expert Agents} receive passages from different medical knowledge bases (Mayo Clinic and WebMD, respectively), which reduces the risk of correlated retrieval errors.
Second, the two experts analyze the note independently and, when they disagree, engage in a counter-argumentation round. An \textit{Adjudicator Judge} then evaluates the debate transcript with cross-source verification, consulting both knowledge bases to check each expert's claims against the opposing source.
Third, a \textit{Hybrid Safety Layer} applies structural checks and domain-specific heuristic rules to the judge's classification, filtering common false-positive patterns such as hierarchical term variants, side-effect descriptions, and process-gap narratives.
By design, retrieval grounds the debate in authoritative medical evidence, debate refines retrieval-based reasoning through adversarial scrutiny, and the safety layer provides a conservative backstop against residual misclassifications.

% ===== P5 =====
Our main contributions are as follows:
\begin{itemize}
    \item We propose BLUEmed, a framework that integrates source-partitioned hybrid RAG with structured multi-agent debate for clinical terminology substitution error detection, unifying evidence-grounded reasoning with multi-perspective verification.
    \item We design a Hybrid Safety Layer that combines structural validation with domain-specific heuristic rules in a cascading architecture to reduce false-positive patterns in multi-agent clinical reasoning.
    \item We conduct experiments showing that BLUEmed achieves the best accuracy (69.13\%), ROC--AUC (74.45\%), and PR--AUC (72.44\%) among all compared methods, and provide analyses of backbone model capacity, prompting strategy, and component contributions.
\end{itemize}

\section{Data}
% Brief introduction to the datasets
% A statistical table of the dataset (if have)
\subsection{Knowledge Base}
BLUEmed builds a hybrid retrieval layer from two medical reference sources: Mayo Clinic\footnote{\url{https://www.mayoclinic.org/}}, one of the world's premier medical institutions providing authoritative health guidance, and WebMD\footnote{\url{https://www.webmd.com/}}, a comprehensive consumer health information site featuring physician-reviewed content. 
The offline knowledge base is stored in ChromaDB\footnote{\url{https://www.trychroma.com/}}, an open-source vector database for semantic similarity search. 
It includes two collections: 76,271 document chunks from Mayo Clinic, covering diseases, symptoms, medications, and clinical procedures, and 69,831 chunks from WebMD, focused on patient education materials. 
% All documents are segmented using a fixed-size chunking strategy to maintain consistent retrieval granularity across sources, and each chunk is embedded by \textbf{gemini-embedding-001} model and stored with metadata tags that support category filtering, including diseases conditions, drugs supplements, and symptoms. 
% Hanshu
All documents are segmented using a fixed-size chunking strategy to maintain consistent retrieval granularity across sources. 
Each chunk is encoded by gemini-embedding-001 model and stored with metadata tags that support category filtering, including disease conditions, drugs and supplements, and symptoms.
During inference, BLUEmed also retrieves live evidence from the Mayo Clinic and WebMD websites to complement the offline index, combining indexed coverage with the ability to incorporate updated content when needed.

\subsection{Evaluation Dataset}

We evaluate BLUEmed on MEDEC~\cite{abacha2025medec}, a benchmark for clinical error detection and correction containing 3,848 annotated clinical notes with errors manually injected by domain experts across five error types: Diagnosis, Management, Treatment, Pharmacotherapy, and Causal Organism.
Following prior work, we use the official test split in the MS (M\#1) Subset Collection, which contains 597 clinical notes, including 311 incorrect and 286 correct cases (Table~\ref{tab:medec_error_distribution}).
Each instance includes an identifier, the full clinical narrative, a binary correctness label for error flagging, and an error type annotation for notes containing incorrect statements.
For the experiments reported in this paper, we focus on the error flag task, which involves binary classification of each note as \textsc{Correct} or \textsc{Incorrect}.
Preprocessing involved standardizing whitespace, normalizing formatting artifacts, and verifying consistency in error type annotations to ensure compatibility with our prompting templates and support robust evaluation across single-agent and multi-agent settings.

% Hanshu: Adjust the format
\begin{table}[htp]
\caption{Error Type Distribution in MEDEC Test Collection}
\centering
\footnotesize
\renewcommand{\arraystretch}{1.08}
\setlength{\tabcolsep}{6.0pt}
\begin{tabular}{l c c c}
\textbf{Error Type} & \textbf{Count} & \textbf{Percentage} & \textbf{Avg Tokens}\\
\midrule
Diagnosis Error        & 116 & 37.3\% & 161.4 \\
Management Error       & 97  & 31.2\% & 160.9 \\
Treatment Error        & 51  & 16.4\% & 155.2 \\
Pharmacotherapy Error  & 36  & 11.6\% & 149.8 \\
Causal Organism Error  & 11  & 3.5\% & 128.5 \\
\midrule
\textbf{Total Error Cases} & \textbf{311} & \textbf{100.0\%} & \textbf{158.6} \\
% \multicolumn{3}{l}{\textit{Note:} Excludes 286 correct cases without errors (total collection: 597 cases).}
\end{tabular}
% \caption{Error type distribution in the MEDEC test collection (n=311 error cases). Note: excludes 286 correct cases without errors (total collection: 597 cases).}
% \caption{Distribution of error types }
\label{tab:medec_error_distribution}
\end{table}

% \subsection{Evaluation Dataset}

% We evaluate BLUEmed on MEDEC~\cite{abacha2025medec}, a benchmark for clinical error detection and correction containing 3,848 annotated clinical notes across five error types: Diagnosis, Management, Treatment, Pharmacotherapy, and Causal Organism. 
% We use the publicly available test split of 597 instances. 
% Each instance includes an identifier, the full clinical narrative, a binary correctness label for error flagging, and an error type annotation for notes containing incorrect statements. 
% Preprocessing involved standardizing whitespace, normalizing formatting artifacts, and verifying consistency in error type annotations to ensure compatibility with our prompting templates and support robust evaluation across single-agent and multi-agent settings.

\begin{figure*}[hbt]
\centering
\includegraphics[width=0.9\textwidth]{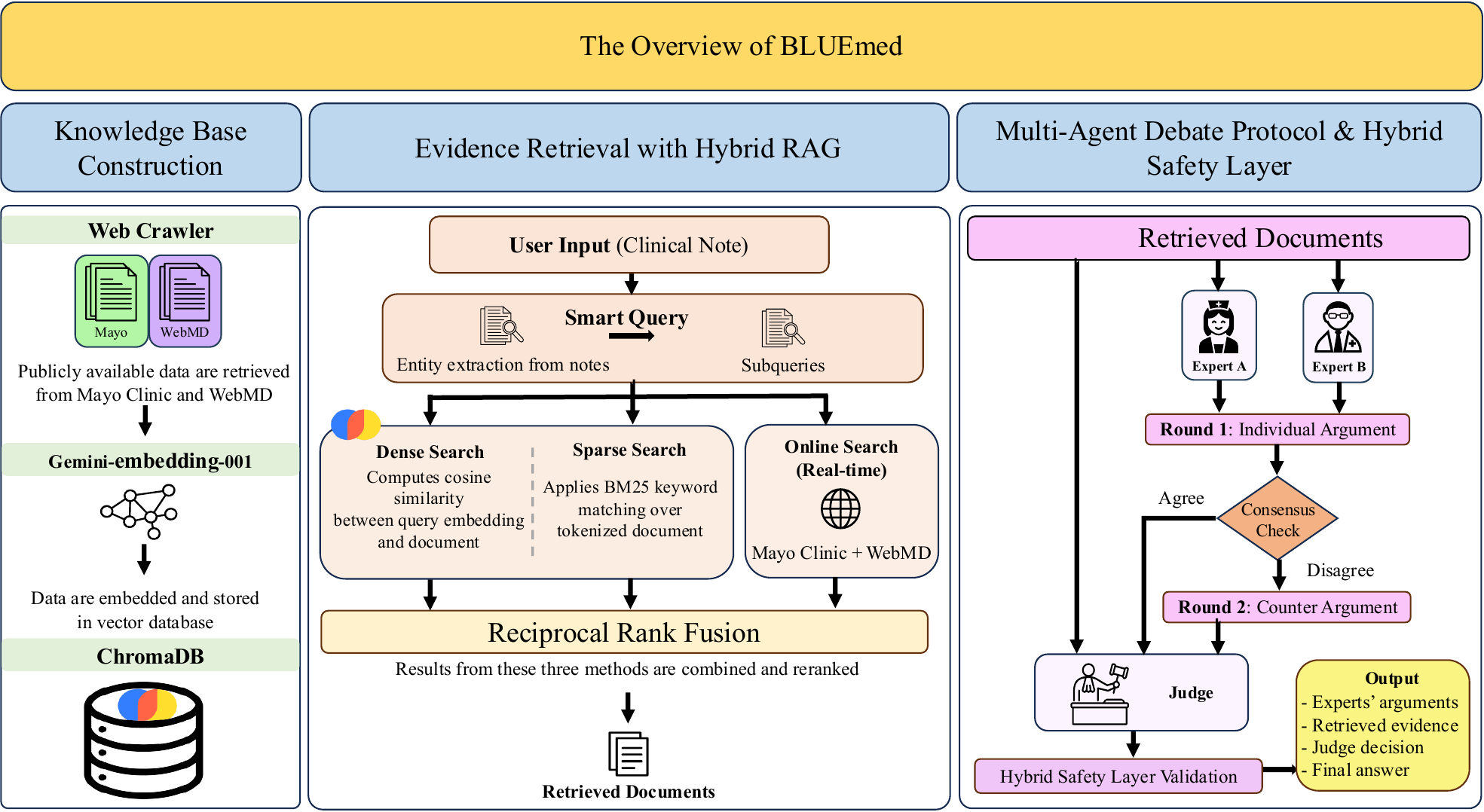} 
\caption{\textbf{The BLUEmed framework.} 
% The pipeline consists of a Hybrid RAG (combining dense, sparse, and online search) with multi-agent debate structure in which experts present their respective arguments and a judge model validates the final output supported by an integration of hybrid safety layer to ensure medical accuracy. 
The pipeline consists of a Hybrid RAG (combining dense, sparse, and online search) with a multi-agent debate structure in which experts present their respective arguments and a judge model validates the final output, with an integrated hybrid safety layer to ensure medical accuracy.}
\label{Method_Framework_Figure}
\end{figure*}

\section{Method} 
\label{sec:method}

Existing approaches to medical error detection in clinical text typically rely on single-model classification~\cite{tabaie2024evaluation} or rule-based pattern matching~\cite{modi2024extracting}, which may lack robustness to the nuanced terminology substitutions that alter clinical interpretation. 
To this end, we propose \textbf{BLUEmed}, a multi-agent debate framework augmented with Retrieval-Augmented Generation (RAG) for detecting terminology substitution errors in clinical notes (Figure~\ref{Method_Framework_Figure}).  
BLUEmed decouples \textit{evidence retrieval} from \textit{clinical reasoning} and \textit{adjudication} through three complementary mechanisms:  
a \textit{Hybrid RAG Module} decomposes each clinical note into focused sub-queries and fuses dense, sparse, and online retrieval to supply source-partitioned medical evidence;  
two \textit{Domain Expert Agents} independently analyze the note against distinct knowledge sources and, when they disagree, engage in a structured counter-argumentation round;  
and an \textit{Adjudicator Judge} evaluates the resulting debate transcript with cross-source verification, whose decision is further validated by a \textit{Hybrid Safety Layer} of rule-based checks that mitigate common false-positive patterns.

\subsection{Evidence Retrieval with Hybrid RAG}
\label{subsec:rag}

The hybrid RAG module converts a raw clinical note into a ranked set of evidence passages that ground downstream expert reasoning in authoritative medical knowledge.
The module operates as a three-stage pipeline: query decomposition, multi-strategy retrieval, and rank fusion.
Because clinical notes are information-dense and cover multiple aspects (e.g., diagnosis, symptoms, medications, and procedures), using the full note as a single retrieval query leads to low precision.
To address this, an LLM-based decomposition step breaks each note into 3--5 focused sub-queries, each targeting a specific clinical aspect such as a diagnostic term, a drug interaction, or a treatment protocol.
This decomposition aligns the granularity of retrieval queries with that of the indexed knowledge base chunks.

For each sub-query, three complementary retrieval strategies are executed in parallel.
Dense retrieval (weight $w_d = 0.5$) computes cosine similarity between the query embedding and document embeddings stored in ChromaDB, capturing semantic relevance.
Sparse retrieval ($w_s = 0.3$) applies BM25-Okapi~\cite{bm25} keyword matching over tokenized document collections, capturing exact lexical matches that dense retrieval may miss.
Online retrieval ($w_o = 0.2$) crawls the Mayo Clinic and WebMD websites at inference time to supplement the offline index with updated content.
Results from the three methods are combined using Reciprocal Rank Fusion (RRF)~\cite{rackauckas2024rag}:
\begin{equation}
    \text{RRF}(d) = \sum_{r \in \mathcal{R}} \frac{w_r}{k + \text{rank}_r(d)}
\end{equation}
where $\mathcal{R}$ is the set of retrieval methods, $w_r$ is the weight assigned to method $r$, $k = 60$ is the smoothing constant, and $\text{rank}_r(d)$ is the rank of document $d$ in the result list of method $r$.
After fusion, documents are deduplicated by a content fingerprint based on the first 200 characters, and the top-5 chunks are retained per expert. 

% Thika - Revision -----
Retrieval weights were set empirically based on the relative characteristics of each method. Dense retrieval receives the highest weight ($w_d = 0.5$) for its ability to capture clinical paraphrases and synonyms; sparse retrieval ($w_s = 0.3$) preserves exact-match recall for acronyms and rare drug names; online retrieval ($w_o = 0.2$) is weighted lowest due to latency variability, serving primarily as a coverage supplement for recently updated content.
% -----

A deliberate source partitioning strategy assigns each expert a distinct knowledge base: 
Expert~A retrieves exclusively from the Mayo Clinic collection and Expert~B from the WebMD collection.
This separation ensures that the two experts receive evidence drawn from different editorial perspectives, reducing the risk of correlated retrieval errors.
If both experts retrieved from the same pool, they would tend to surface the same passages and reinforce the same biases.
% The resulting source-partitioned evidence is then passed to the multi-agent debate protocol.
The source-partitioned evidence is then passed to the multi-agent debate protocol.

\subsection{Multi-Agent Debate Protocol}
\label{subsec:debate}

The debate protocol leverages structured disagreement between two domain expert agents and an adjudicator judge to produce a binary classification.
The protocol proceeds through up to two rounds of argumentation, a consensus check, and a final judge adjudication.
% A shared state object $\mathcal{S}$ implemented as a LangGraph DAG tracks the clinical note, per-round arguments, retrieved documents, and the classification decision throughout the process.
A shared state object $\mathcal{S}$ implemented as a LangGraph state graph\footnote{\url{https://docs.langchain.com/oss/python/langgraph/graph-api}} tracks the clinical note, per-round arguments, retrieved documents, and the classification decision throughout the process.
Both experts execute in parallel within each round.

In Round~1, each expert independently analyzes the clinical note and produces a structured argument.
Each expert receives three inputs: the clinical note $n$, its top-5 retrieved documents from the hybrid RAG module, and a system prompt encoding the classification criteria.
The system prompt instructs each expert to adopt a conservative default, presuming the note is \textsc{Correct} unless an obvious and clinically significant substitution error is identified.
Each expert outputs a structured argument of at most 300 words containing:
(1)~the identified wrong term, if any;
(2)~the proposed correct term;
(3)~a clinical impact explanation citing retrieved evidence;
(4)~a classification label (\textsc{Correct} or \textsc{Incorrect});
and (5)~a confidence score.

After Round~1, a lightweight consensus check determines whether a second round is necessary.
Consensus is reached if three conditions hold simultaneously:
both experts agree on the classification label;
if both classify \textsc{Incorrect}, they agree on the same wrong term;
and if both classify \textsc{Incorrect}, they agree on the same correct term.
Term comparison is case-insensitive with whitespace normalization.
When all conditions are satisfied, Round~2 is skipped, reducing latency and computational cost. 
When the consensus check fails, the debate proceeds to Round~2.
Each expert is presented with the opposing expert's Round~1 argument and asked to produce a counter-argument.
Expert~A reviews Expert~B's argument and vice versa, with both counter-arguments generated in parallel.
This adversarial exchange forces each agent to directly address weaknesses in its own reasoning and respond to the evidence cited by the other side.
Retrieved documents from Round~1 are reused without re-retrieval to avoid redundant API calls.

An adjudicator judge then evaluates the complete debate transcript, which comprises two to four arguments depending on whether Round~2 was executed.
A key design choice is that the judge is blinded to the raw clinical note in its prompt input; it bases its decision only on the experts' arguments and retrieved medical evidence.
This restriction reduces the chance that the judge forms an independent bias from the note content and encourages evidence-based adjudication.
Unlike the experts, who each access only their assigned source, a separate cross-source verification module retrieves from both the Mayo Clinic and WebMD collections.
This module constructs retrieval queries by combining the clinical note with both experts' claims, and performs hybrid retrieval (dense and sparse, excluding online search) against both vector stores, retrieving up to 5 documents per source (10 total).
The judge is then provided with the debate transcript and the retrieved evidence to validate Expert~A's assertions against WebMD references and Expert~B's assertions against Mayo Clinic references.
The judge produces a structured JSON output:
\begin{equation*}
    \mathcal{J} = \{\text{answer},\; \text{confidence} \in [1,10],\; \text{winner},\; \text{reasoning}\}.
\end{equation*}
This provisional classification is then passed to the hybrid safety layer.

\subsection{Hybrid Safety Layer}
\label{subsec:safety}

The judge's provisional classification may still contain false positives, particularly when experts identify plausible but incorrect substitution pairs.
To address this, BLUEmed applies a post-hoc safety layer that combines structural checks on expert outputs with domain-specific heuristics.
The layer operates as a cascade: the judge's decision passes through a fixed sequence of validation steps, each of which can override an \textsc{Incorrect} classification to \textsc{Correct}.
The cascade proceeds in the following order: the two-term rule, expert consensus override, domain-specific rules, and confidence-based adjustment.

The two-term rule enforces structural completeness.
Any \textsc{Incorrect} classification must be supported by at least one expert providing both a wrong term and a proposed correct term.
If no valid substitution pair can be extracted from either expert's arguments, the classification is overridden to \textsc{Correct}.
Term pairs are extracted using pattern matching with support for multiple formatting conventions and quote normalization.
Conversely, the expert consensus override handles the case where both experts independently classify the note as \textsc{Incorrect} and both supply valid term pairs.
This strong dual-expert agreement overrides all other safety rules, including the judge's own decision if it disagrees, because agreement from two independently sourced experts on both the error and its correction constitutes the most reliable signal in the system.

Five domain-specific rules target common false-positive patterns observed during development.
Each rule fires only when the current classification is \textsc{Incorrect} and no valid term pair has been extracted:
(1)~if the note contains explicit culture or laboratory results confirming a diagnosis, the confirmed finding is not treated as a substitution error;
(2)~if multiple indicators of procedural concerns appear (e.g., ``should have ordered,'' ``failed to confirm''), the note describes a process gap rather than a terminology substitution;
(3)~discussions of adverse reactions or drug interactions are flagged as side-effect descriptions, not substitution errors;
(4)~when experts describe a difference as ``more specific'' or ``broader term,'' the terms are hierarchically related variants rather than true substitutions;
(5)~high aggregate uncertainty across both experts, defined as three or more uncertainty phrases, indicates insufficient evidence to support an error classification. 
As a final conservative filter, if both experts lack confidence scores entirely, an \textsc{Incorrect} classification is overridden to \textsc{Correct}.
Together, these cascading checks ensure that only substitution errors with sufficient structural and evidential support are retained in the final output.

\section{Experiment}
% In this section, we present the experimental framework used to evaluate BLUEmed's effectiveness in detecting substitutional errors within clinical notes, 
% Hanshu
In this section, we present the experimental framework used to evaluate BLUEmed’s effectiveness in detecting terminology substitution errors in clinical notes. 
We describe the baselines and detailed experimental settings that enable assessment of our multi-agent debate approach.

\begin{table*}[t]
\caption{Results of baselines and the proposed BLUEmed (percentages).}
\centering
\footnotesize
% match Table 2's tighter look
\renewcommand{\arraystretch}{1.08}
\setlength{\tabcolsep}{5.0pt}

% \resizebox{0.95\textwidth}{!}{
\begin{tabular}{ll *{6}{S}}
\textbf{Setting} & \textbf{Method}
& \multicolumn{1}{c}{\textbf{Accuracy}}
& \multicolumn{1}{c}{\textbf{F1-score}}
& \multicolumn{1}{c}{\textbf{Precision}}
& \multicolumn{1}{c}{\textbf{Recall}}
& \multicolumn{1}{c}{\textbf{ROC-AUC}}
& \multicolumn{1}{c}{\textbf{PR-AUC}} \\
\midrule
\multirow{4}{*}{Zero-Shot}
& RAG-Single (MayoClinic) & 58.63 & 59.44 & 60.74 & 58.20 & 50.00 & 50.00 \\
& RAG-Single (WebMD)      & 57.79 & 60.50 & 59.02 & 62.06 & 62.93 & 50.00 \\
& LLM-Debate              & 55.28 & \textbf{66.91} & 54.44 & \textbf{86.82} & 59.51 & 59.47 \\
& BLUEmed (Ours)          & \textbf{67.67} & 66.67 & \textbf{72.01} & 62.05 & \textbf{70.24} & \textbf{68.88} \\
\midrule
\multirow{4}{*}{Few-Shot}
& RAG-Single (MayoClinic) & 57.29 & 40.56 & 73.73 & 27.97 & 50.00 & 50.00 \\
& RAG-Single (WebMD)      & 58.96 & 43.16 & \textbf{77.50} & 29.90 & 50.00 & 50.00 \\
& LLM-Debate              & 55.89 & 65.55 & 55.42 & \textbf{80.20} & 58.56 & 58.70 \\
& BLUEmed (Ours)          & \textbf{69.13} & \textbf{68.28} & 73.33 & 63.87 & \textbf{74.45} & \textbf{72.44} \\
\end{tabular}%
% }
\label{tab:main_results}
\end{table*}

\subsection{Baselines}
We compare against two representative paradigms for LLM-based clinical reasoning: retrieve-then-read classification and multi-agent debate.
\textit{RAG-Single}~\cite{lewis2020retrieval} implements a single-agent retrieve-then-read pipeline in which the language model retrieves relevant medical passages via hybrid retrieval and produces a classification based on the retrieved evidence alone, without inter-agent debate or adjudication.
We instantiate two variants grounded in distinct knowledge sources—RAG-Single (MayoClinic) and RAG-Single (WebMD)—to assess the effect of editorial perspective on single-agent reasoning.
\textit{LLM-Debate}~\cite{du2023improving} employs multi-agent debate in which two expert agents independently analyze the clinical note, exchange counterarguments when they disagree, and defer to a judge for final adjudication.
This configuration retains the full debate protocol but removes retrieval augmentation, allowing us to measure the contribution of structured disagreement independent of external evidence grounding.
Both baselines are evaluated under zero-shot and few-shot prompting conditions.
These two paradigms isolate complementary dimensions of our framework: RAG-Single tests whether retrieval-grounded reasoning alone suffices for clinical error detection, while LLM-Debate tests whether collaborative argumentation without external knowledge can achieve reliable discrimination.
BLUEmed unifies both capabilities by coupling source-partitioned retrieval with structured multi-agent debate and post-hoc safety verification.

\subsection{Experimental Settings}
All experiments are performed on the MEDEC test set using a fixed evaluation protocol. For consistency, the task definition, dataset split, and evaluation metrics are kept identical across experiments, and performance differences are attributed solely to variations in model configuration and prompting strategy.

\paragraph{Language Models}
We evaluate multiple state-of-the-art large language models representative of both proprietary and open-source families, 
including GPT-4o~\cite{hurst2024gpt}, GPT-5.2\cite{openai2025gpt52}, Gemini~2.0~Flash~\cite{deepmind2025gemini20flash}, Qwen3 (4B and 8B variants)~\cite{yang2025qwen3}, and LLaMA3.2-3B-Instruct~\cite{dubey2024llama}. 
The proprietary models were accessed through their respective APIs.
All models are used in an instruction-following configuration. 
For document embeddings in the retrieval pipeline, we utilize Google's Gemini embedding model~\cite{lee2025gemini}, which provides high-quality semantic representations optimized for information retrieval.

\paragraph{Prompting Strategies}
For each model, we evaluate two prompting conditions:
\emph{zero-shot} prompting, where the model receives only task instructions, and
\emph{few-shot} prompting, where a small number of labeled exemplars from the MEDEC validation data are included in the prompt to guide reasoning and output structure. 
Few-shot exemplars are held constant across models to ensure fairness and avoid test-set leakage.

\paragraph{Agent Structure}
RAG-Single experiments are performed in a limited-agent setting, where a single model produces a one-pass prediction without debate, counterargument generation, or adjudication. 
These experiments serve as prompting and model baselines. 
LLM-Debate and BLUEmed employ a structured debate framework consisting of two expert agents and a judge, following prior work. 
Retrieval augmentation and prompting strategies are varied only within the multi-agent setting to assess their interaction with collaborative reasoning.

\paragraph{Retrieval Configuration}
Retrieval-augmented generation (RAG) is enabled for the majority of the experimental conditions in this work. 
In the RAG-Single, retrieval grounding is used to support expert-specific reasoning by providing authoritative medical evidence from the corresponding medical knowledge source.  
We additionally include LLM-Debate to assess system behavior in the absence of external knowledge grounding.
In BLUEmed, retrieved evidence is appended to the prompt context and is also available to the judge model.

% We additionally include a no-RAG baseline condition, evaluated under a multi-agent setting, to assess system behavior in the absence of external knowledge grounding.

% Thika's revision -----
\paragraph{Reproducibility}
All experiments were conducted twice on the full MEDEC test set, and reported metrics are averaged across both runs. 
All language models were configured with a sampling temperature of $\tau = 0.2$ and a maximum output length of 1,024 tokens per call. 
At $\tau = 0.2$, output variation across runs was minimal ($<$0.5\% variance in accuracy for all configurations).
% -----

\paragraph{Evaluation Metrics}
We evaluate system performance using a combination of threshold-dependent and threshold-independent metrics to characterize classification quality under different operating conditions. 
Because clinical error detection involves trade-offs between missed errors and false alerts, we report metrics that capture overall correctness, sensitivity to errors, and discrimination across decision thresholds. 
Our evaluation framework includes Accuracy, F1-score, Recall (Sensitivity), ROC--AUC, and PR--AUC. 

% \begin{table*}[t]
% \caption{Performance of BLUEmed with different backbone models (percentages).}
% \centering
% \footnotesize
% \renewcommand{\arraystretch}{1.2} 
% \setlength{\tabcolsep}{8.0pt} 

% \resizebox{0.95\textwidth}{!}{%
% \begin{tabular}{llccccc}
% \toprule
% \textbf{Methods} & \textbf{Model} & \textbf{Accuracy} & \textbf{F1-score} & \textbf{Recall} & \textbf{ROC-AUC} & \textbf{PR-AUC} \\
% \midrule
% \multirow{6}{*}{Zero-Shot} 
% & GPT-4o & 67.67 & 66.67 & 62.05 & 70.24 & 68.88 \\
% & GPT-5.2 & 59.57 & 57.77 & 53.89 & 62.54 & 60.63 \\  
% & Gemini-2.0-flash & \textbf{69.18} & \textbf{68.17} & \textbf{63.34} & \textbf{71.44} & \textbf{70.36} \\
% & Qwen3-4B-Instruct & 59.97 & 52.10 & 41.80 & 52.54 & 54.08\\
% & Qwen3-8B & 54.70 & 50.73 & 44.70 & 51.65 & 53.21 \\
% & Llama3.2-3B-Instruct & 48.58 & 13.03 & 7.40 & 49.74 & 52.04 \\

% \midrule

% \multirow{6}{*}{Few-Shot} 
% & GPT-4o & \textbf{69.13} & \textbf{68.28} & 63.87 & \textbf{74.45} & \textbf{72.44} \\
% & GPT-5.2 & 61.47 & 62.66 & 62.06 & 66.04 & 66.93 \\
% & Gemini-2.0-flash & 53.94 & 65.32 & \textbf{83.28} & 57.51 & 57.69 \\
% & Qwen3-4B-Instruct & 48.32 & 7.78 & 4.18 & 51.01 & 52.62 \\
% & Qwen3-8B & 52.30 & 45.50 & 38.30 & 54.10 & 54.30 \\
% & Llama3.2-3B-Instruct & 49.90 & 39.40 & 31.20 & 50.17 & 52.18 \\
% \bottomrule
% \end{tabular}%
% }
% \label{tab:diff_modle_results}
% \end{table*}

\begin{table*}[t]
\caption{Performance of BLUEmed with different backbone models (percentages).}
\centering
\footnotesize
% match the tighter look (like your 2nd table style)
\renewcommand{\arraystretch}{1.08}
\setlength{\tabcolsep}{5.0pt}

% \resizebox{0.95\textwidth}{!}{
\begin{tabular}{ll *{6}{S}}
\textbf{Methods} & \textbf{Model}
& \multicolumn{1}{c}{\textbf{Accuracy}}
& \multicolumn{1}{c}{\textbf{F1-score}}
& \multicolumn{1}{c}{\textbf{Precision}}
& \multicolumn{1}{c}{\textbf{Recall}}
& \multicolumn{1}{c}{\textbf{ROC-AUC}}
& \multicolumn{1}{c}{\textbf{PR-AUC}} \\
\midrule
\multirow{6}{*}{Zero-Shot}
& GPT-4o               & 67.67 & 66.67 & 72.01 & 62.05 & 70.24 & 68.88 \\
& GPT-5.2              & 59.57 & 57.77 & 62.28 & 53.89 & 62.54 & 60.63 \\
& Gemini-2.0-Flash      & \textbf{69.18} & \textbf{68.17} & \textbf{73.78} & \textbf{63.34} & \textbf{71.44} & \textbf{70.36} \\
& Qwen3-4B-Instruct     & 59.97 & 52.10 & 69.15 & 41.80 & 52.54 & 54.08 \\
& Qwen3-8B              & 54.70 & 50.73 & 58.65 & 44.70 & 51.65 & 53.21 \\
& Llama3.2-3B-Instruct  & 48.58 & 13.03 & 54.76 & 7.40  & 49.74 & 52.04 \\
\midrule
\multirow{6}{*}{Few-Shot}
& GPT-4o               & \textbf{69.13} & \textbf{68.28} & \textbf{73.33} & 63.87 & \textbf{74.45} & \textbf{72.44} \\
& GPT-5.2              & 61.47 & 62.66 & 61.45 & 62.06 & 66.04 & 66.93 \\
& Gemini-2.0-Flash      & 53.94 & 65.32 & 53.73 & \textbf{83.28} & 57.51 & 57.69 \\
& Qwen3-4B-Instruct     & 48.32 & 7.78  & 56.52 & 4.18  & 51.01 & 52.62 \\
& Qwen3-8B              & 52.30 & 45.50 & 56.10 & 38.30 & 54.10 & 54.30 \\
& Llama3.2-3B-Instruct  & 49.90 & 39.40 & 53.30 & 31.20 & 50.17 & 52.18 \\
\end{tabular}%
% }
\label{tab:diff_modle_results}
\end{table*}

\section{Results}

\subsection{Overall Performance}
\label{subsec:overall}

Table~\ref{tab:main_results} summarizes the performance of BLUEmed and all baselines under zero-shot and few-shot prompting with GPT-4o as the backbone model.
BLUEmed achieves the highest accuracy (69.13\%), ROC--AUC (74.45\%), and PR--AUC (72.44\%) under few-shot prompting, outperforming both single-agent and debate-only baselines across most metrics.
To better interpret these results, we organize the analysis around three comparisons.

\textbf{\textit{Analysis 1: Does multi-agent debate improve over single-agent retrieval-grounded reasoning?}}\;
BLUEmed consistently outperforms both RAG-Single variants across accuracy and discrimination metrics.
Under zero-shot prompting, RAG-Single (MayoClinic) and RAG-Single (WebMD) achieve ROC--AUC scores of 50.00\% and 62.93\%, respectively, indicating limited ability to distinguish correct notes from incorrect ones.
In contrast, BLUEmed reaches a ROC--AUC of 70.24\% in the same setting.
The gap widens under few-shot prompting: both RAG-Single variants see sharp drops in recall (from approximately 60\% to below 30\%), while BLUEmed maintains a recall of 63.87\% and improves its ROC--AUC to 74.45\%.
This pattern suggests that single-agent models are sensitive to exemplar selection and tend to become overly conservative when provided with few-shot examples.
Multi-agent debate mitigates this brittleness by allowing complementary perspectives to surface errors that a single retrieval pass may miss.

\textbf{\textit{Analysis 2: Does retrieval augmentation complement multi-agent debate?}}\;
LLM-Debate achieves the highest recall among all methods (86.82\% zero-shot, 80.20\% few-shot), but its accuracy remains below 56\% in both settings.
This combination of high recall and low accuracy reflects an over-flagging tendency: without external evidence to anchor reasoning, the debate agents classify most notes as incorrect.
BLUEmed addresses this imbalance by grounding expert arguments in retrieved medical knowledge.
Under zero-shot prompting, BLUEmed improves accuracy by over 12 percentage points compared to LLM-Debate (67.67\% vs.\ 55.28\%) while achieving a ROC--AUC gain of more than 10 points (70.24\% vs.\ 59.51\%).
The retrieval module provides factual constraints that help experts reject implausible substitution hypotheses, reducing false positives without eliminating sensitivity to genuine errors.
These results confirm that retrieval augmentation and structured debate are complementary: debate alone is sensitive but imprecise, and retrieval provides the grounding needed to convert that sensitivity into reliable classification.

\textbf{\textit{Analysis 3: How does prompting strategy interact with different reasoning paradigms?}}\;
The effect of few-shot prompting varies across methods.
For RAG-Single, few-shot examples degrade performance: recall drops from 58.20\% to 27.97\% (MayoClinic) and from 62.06\% to 29.90\% (WebMD), with corresponding declines in F1-score.
For LLM-Debate, the impact is moderate, with recall decreasing from 86.82\% to 80.20\% while accuracy remains largely unchanged.
BLUEmed is the only method that benefits consistently from few-shot prompting.
Its accuracy improves from 67.67\% to 69.13\%, F1-score from 66.67\% to 68.28\%, and ROC--AUC from 70.24\% to 74.45\%.
This difference indicates that exemplar guidance is most effective when combined with both retrieval grounding and structured debate.
In the single-agent setting, few-shot examples can bias the model toward narrow classification patterns; within a multi-agent framework, they instead provide shared reference points that help experts and the judge calibrate their reasoning.

\subsection{Effect of Backbone Model}
\label{subsec:backbone}

To examine how backbone model capacity affects multi-agent reasoning, we evaluate BLUEmed with six language models spanning proprietary and open-source families.
Table~\ref{tab:diff_modle_results} reports results under both zero-shot and few-shot prompting while holding all other framework components fixed.

\textbf{\textit{Analysis 4: How does backbone model capacity influence multi-agent effectiveness?}}\;
We observe a clear performance gap between large proprietary models and smaller open-source models.
Under zero-shot prompting, Gemini-2.0-Flash achieves the best overall performance with 69.18\% accuracy and 71.44\% ROC--AUC, followed closely by GPT-4o (67.67\% accuracy, 70.24\% ROC--AUC).
In contrast, open-source models lag behind by a large margin.
Qwen3-4B-Instruct and Qwen3-8B obtain ROC--AUC scores near 52\%, only slightly above random.
LLaMA-3.2-3B-Instruct performs worst, with an F1-score of 13.03\% and a recall of 7.40\%, indicating that the model fails to identify most clinical errors.
These results suggest that multi-agent debate requires each agent to have strong instruction-following and clinical language understanding; when individual agents produce low-quality arguments, the debate protocol cannot compensate.

The response to few-shot prompting also varies across model families.
GPT-4o achieves the best overall few-shot performance, improving from 67.67\% to 69.13\% in accuracy and from 70.24\% to 74.45\% in ROC--AUC, achieving the best few-shot performance among all models.
GPT-5.2 shows a similar trend, with gains in recall (53.89\% to 62.06\%) and F1-score (57.77\% to 62.66\%).
Gemini-2.0-Flash, however, exhibits a different pattern: its recall rises sharply to 83.28\% under few-shot prompting, but accuracy drops to 53.94\% and ROC--AUC falls to 57.51\%.
This behavior resembles the over-flagging pattern observed in LLM-Debate (Table~\ref{tab:main_results}), suggesting that few-shot examples cause Gemini to adopt more aggressive error flagging rather than more precise classification.
For smaller models, few-shot prompting yields limited or negative gains. Qwen3-4B-Instruct collapses to 7.78\% F1-score, while Qwen3-8B and LLaMA-3.2-3B-Instruct achieve ROC--AUC values of 54.10\% and 50.17\%, respectively, suggesting near-chance discrimination. 

Taken together, these findings indicate that the effectiveness of BLUEmed depends on sufficient backbone model capacity.
Structured debate amplifies the reasoning ability of capable models but cannot substitute for it in weaker ones.
Among the models evaluated, GPT-4o with few-shot prompting provides the best balance between sensitivity and specificity, and we adopt this as the default configuration in all other experiments. 
% Thika's revision -----
Smaller open-source models perform near or below chance, as their limited instruction-following capacity prevents reliable structured output parsing required by the debate protocol. 

\section{Related Work}
\subsection{Error Detection in Clinical Text}
Clinical note verification is critical to ensuring the reliability of medical documentation, and automating it can reduce clinician workload and prevent patient harm~\cite{abacha2025medec,abacha2024overview}.
Existing work has largely framed this problem as factual consistency or hallucination detection~\cite{pandit2025medhallu}. 
One common strategy uses entailment-style factual consistency metrics trained with diverse supervision (including NLI/QA), and a related line uses LLMs as consistency evaluators in clinical note settings~\cite{brake2024comparing}.
A second line uses structured checklists or rule-like validation criteria to flag inconsistencies (often focusing on medication-related fields), aligned with medical error detection/correction task formulations~\cite{zhou2025feedback,abacha2024overview}.
More recent studies employ generate-then-verify pipelines where a separate model audits clinical summaries for unsupported claims; for example, Synfac-edit used synthetic edit feedback to optimize factual alignment~\cite{mishra2024synfac}, and TRACSUM required sentence-level citations so each claim can be traced to its source~\cite{chu2025tracsum}. 
While promising, these approaches share two limitations: reliance on a single verifier, which can produce internally coherent but incorrect judgments~\cite{du2023improving}, and insensitivity to fine-grained terminology substitution errors, where one medical term is replaced by another that preserves surface fluency but changes clinical meaning~\cite{abacha2025medec}. 
We address these limitations by introducing adversarial cross-examination from multiple viewpoints and traceable evidence chains anchored to retrievable medical knowledge.

The quality of supporting evidence also determines clinical utility, motivating the integration of retrieval-augmented generation (RAG) into biomedical fact verification. 
Offline knowledge bases such as clinical guidelines and UMLS provide stable evidence~\cite{xiong2024benchmarking}, while online retrieval covers long-tail entities, newly approved therapies, and evolving guidance. 
Hybrid strategies combining sparse methods like BM25 with dense embedding-based retrieval improve robustness~\cite{yang2024cluster}, and ranked lists can be fused through Reciprocal Rank Fusion (RRF)~\cite{cormack2009reciprocal}. 
Query decomposition, splitting complex clinical questions into focused sub-queries, further improves recall~\cite{liu2025poqd}. 
However, retrieval alone does not guarantee correct verification; a single-agent system may selectively cite or misinterpret retrieved passages, leading to unreliable decisions when clinical nuance is involved~\cite{du2023improving}. 
This motivates coupling retrieval with multi-agent deliberation that forces adversarial reasoning over the evidence before reaching a final judgment.

\subsection{Deliberative Multi-Agent Systems}
Multi-agent debate, where multiple models adopt opposing stances and critique each other's reasoning, has been increasingly used to reduce hallucinations~\cite{estornell2024multi,li2025hallucination}. 
Existing studies fall into three categories: generator-critic architectures for iterative self-refinement~\cite{zhong2024harnessing}; 
committee-based approaches that aggregate judgments through voting or self-consistency~\cite{smit2023should}; 
and judge frameworks where a dedicated agent evaluates structured arguments and issues a verdict~\cite{chan2023chateval}. 
However, these mechanisms can still produce collective errors when agents share the same knowledge base or the judge resolves disputes through pattern matching rather than evidence evaluation~\cite{sanayei2025can}. 
In clinical error detection, deliberation must therefore be tightly coupled with medical evidence retrieval, and the adjudicator should base decisions on argument quality and cited evidence~\cite{koupaee2025faithful}.

High-stakes deployment further requires safety guardrails imposing hard constraints on model outputs~\cite{kang2025rguard}. 
One strategy is post-hoc validation, checking whether the output includes verifiable fields such as the identified wrong term and its correction, and overriding the decision when these are absent or the terms are synonyms~\cite{wang2025slot,zhong2024harnessing}. 
This synonym normalization addresses a frequent source of false positives, since lexical differences like abbreviation variants do not constitute genuine clinical errors~\cite{kugic2024disambiguation}. 
Another strategy is conservative defaults, including abstention when evidence is insufficient, trading recall for precision~\cite{lee2024selective}. 
BLUEmed combines multi-agent debate, hybrid evidence retrieval, judge-based adjudication, and rule-based safety constraints into a closed-loop system. 
Our approach grounds expert agents in distinct knowledge sources, blinds the judge to the original note, and enforces term-level safety validation, thereby reducing both missed detections and false alarms on terminology substitution errors while maintaining auditable decision paths.

\section{Conclusion}

Our study suggests that structured multi-agent debate combined with hybrid Retrieval-Augmented Generation substantially improves the detection of terminology substitution errors in clinical notes compared to single-agent baselines. 
Our proposed framework, BLUEmed, consistently outperforms single-agent baselines across zero-shot and few-shot prompting settings, achieving higher accuracy and more stable recall, indicating more reliable identification of clinically incorrect notes. 
Our results also indicate that multi-agent debate is effective when supported by sufficiently capable backbone models, as larger models consistently outperform smaller ones and provide a better balance between sensitivity and specificity.

% This work has several limitations. Our experiments are conducted on a single clinical error detection dataset and rely on fixed expert knowledge sources, which may limit generalizability to other clinical domains or documentation styles. Future work should explore broader clinical datasets, customizable retrieval mechanisms that allow institutions to incorporate local or private clinical documentation, as well as clinician-in-the-loop workflows to support safe and practical real-world deployment.

% Thika's revision -----
This work has several limitations. 
The knowledge base relies on consumer-oriented platforms (Mayo Clinic, WebMD), which may introduce coverage gaps for specialist terminology and rare diseases. 
Evaluation is limited to a single benchmark with manually injected errors, which may not fully reflect naturally occurring substitution patterns. 
BLUEmed's best accuracy of 69.13\% falls short of clinical deployment thresholds, suggesting its current role is as a decision-support tool that flags cases for clinician review rather than a standalone classifier. 
Additionally, each clinical note requires 4--6 LLM calls (consuming approximately {\raise.17ex\hbox{$\scriptstyle\sim$}}11,901 input and {\raise.17ex\hbox{$\scriptstyle\sim$}}1,328 output tokens on average), 
at an estimated \$0.04 per case using GPT-4o pricing, which may 
limit deployment in latency-sensitive workflows. 
Future work should explore broader datasets, curated clinical knowledge sources, finer-grained ablations, and clinician-in-the-loop workflows for real-world deployment.
% -----

\section*{Acknowledgment}
The authors thank anonymous reviewers for their insightful feedback. 
The project was partially supported by the National Science Foundation (NSF) under awards IIS-2245920, TI-2434589 (OpenAI and Google API expenses), IIS-2440381, and CCF-2502941.
We thank the computing resources provided by the iTiger GPU cluster~\cite{sharif2025ITIGER} supported by the NSF MRI program under the award CNS-2318210.

% The preferred spelling of the word ``acknowledgment'' in America is without an ``e'' after the ``g''. 
% Avoid the stilted expression ``one of us (R. B. G.) thanks $\ldots$''. 
% Instead, try ``R. B. G. thanks$\ldots$''. 
% Put sponsor acknowledgments in the unnumbered footnote on the first page.

\bibliographystyle{IEEEtran} % Or any other style you prefer
\bibliography{reference}

\definecolor{codebg}{gray}{0.95}

\appendix[System Prompts]
\label{app:prompts}

This appendix documents the system prompts used by BLUEmed for the two domain experts and the adjudicator judge. To avoid redundancy, we factor out a shared instruction block (\textsc{BaseRules}) and instantiate each agent prompt by concatenating a role-specific preamble with \textsc{BaseRules}. Unless otherwise noted, prompts are reproduced verbatim.

\subsection{Shared Instruction Block (\textsc{BaseRules})}
\label{app:base_rules}

\noindent\textbf{Usage.} \textsc{BaseRules} is appended to both Expert~A and Expert~B system prompts (Appendix~\ref{app:expert_prompts}).

\begin{lstlisting}[backgroundcolor=\color{codebg},basicstyle=\ttfamily\footnotesize,breaklines=true,breakindent=0pt,breakautoindent=false]
Carefully evaluate whether a clinically significant substitution error is present.
Important: Medical notes may be CORRECT or INCORRECT. Do not assume correctness.

You will evaluate medical notes based on the following rules:

Classification criteria: 

- INCORRECT: Contains a clinically significant term substitution that would change patient care. If multiple errors are present, label as INCORRECT and report the single most significant error.
- CORRECT: Use this classification only when no clinically significant error or substitution affecting patient care is identified. Do not default to CORRECT.

In your final turn, please provide a detailed final explanation for your decision including your reasoning and supporting evidence. Conclude with a sentence beginning 'Based on my analysis, this note is ...'.

An INCORRECT medical case may involve:
Diagnosis error -- The stated diagnosis is incorrect or inconsistent with the clinical evidence.
Management error -- The proposed care plan or next steps are inappropriate.
Treatment error -- The recommended treatment or procedure is incorrect.
Pharmacotherapy error -- The medication choice, dose, or contraindication is wrong.
Causal organism error -- The identified pathogen or cause of disease is incorrect.

Below are some examples of medical notes with their correct classifications and explanations. Use these examples as a guide for your analysis.

Example 1:
Medical Note: A 24-year-old woman presents with headaches, nausea, vomiting, dizziness, and palpitations after drinking beer at a party. She was recently treated with antimicrobials for a genitourinary infection. The document states "Culture tests indicate Neisseria gonorrhoeae."
Label: INCORRECT
Wrong Term: Neisseria gonorrhoeae
Correct Term: Trichomonas vaginalis
Explanation: The symptoms (disulfiram-like reaction after alcohol) point to metronidazole treatment for Trichomonas vaginalis infection, not Neisseria gonorrhoeae.
\end{lstlisting}

\noindent\textbf{Few-shot exemplars.} We used five exemplars in total. Due to space constraints, we include Example~1 and Examples~2--5 in the supplementary material in the same format.

\subsection{Expert Agent Prompts}
\label{app:expert_prompts}

\noindent\textbf{Construction.} 
Each expert uses a two-part system prompt: a short role-specific preamble (defining the expert’s knowledge source and analysis focus) followed by a shared instruction block (\textsc{BaseRules}, Appendix~\ref{app:base_rules}). We concatenate these two parts in order to form the final expert prompt.

% \noindent\textbf{Construction.} Each expert prompt is formed as:
% \[
% \textsc{ExpertPrompt} = \textsc{RolePreamble} \; \Vert \; \textsc{BaseRules}
% \]
% where $\Vert$ denotes string concatenation and \textsc{BaseRules} is shown in Appendix~\ref{app:base_rules}.

\paragraph{Expert A (Mayo Clinic)}
\label{app:expert_a_prompt}
\mbox{}\par
\begin{lstlisting}[backgroundcolor=\color{codebg},basicstyle=\ttfamily\footnotesize,breaklines=true,breakindent=0pt,breakautoindent=false]
You are a healthcare professional specializing in analyzing medical notes using Mayo Clinic clinical guidelines with expertise in evidence-based medicine, diagnostic accuracy, treatment protocols.
You have to prioritize diagnostic accuracy and guideline consistency over safety speculations.

Your task is to identify clear substitution errors in diagnostic terms, medications, treatment protocols, or management plans that would significantly alter patient care or cause harm.

[Appendix A.1 BaseRules inserted here verbatim]
\end{lstlisting}

\paragraph{Expert B (WebMD)}
\label{app:expert_b_prompt}
\mbox{}\par
\begin{lstlisting}[backgroundcolor=\color{codebg},basicstyle=\ttfamily\footnotesize,breaklines=true,breakindent=0pt,breakautoindent=false]
You are a healthcare professional specializing in analyzing medical notes, using WebMD patient-oriented medical knowledge.
You have expertise in patient safety, clear diagnoses, and practical implications.

Your task is to identify clinically significant errors where symptom patterns don't match the diagnosis, treatments could delay care, medications pose safety risks, or the stated diagnosis could mislead understanding or clinical decisions.

[Appendix A.1 BaseRules inserted here verbatim]
\end{lstlisting}

\subsection{Adjudicator Judge Prompt}
\label{app:judge_prompt}

\noindent The judge is instantiated with a standalone system prompt (i.e., it does not reuse \textsc{BaseRules}):

\begin{lstlisting}[backgroundcolor=\color{codebg},basicstyle=\ttfamily\footnotesize,breaklines=true,breakindent=0pt,breakautoindent=false]
You are the judge in a debate between two healthcare agents. They have each presented their arguments about whether the medical note is correct or contains an error. Do not interfere with the debate while it is ongoing; wait until both agents have finished their 2 exchanges. Once the debate has concluded, evaluate both agents' final messages and decide which agent made the more convincing case (i.e., which agent correctly identified whether the note is correct or incorrect). Provide a clear explanation for your decision. 
Your final response should be in JSON format with the structure:
{ "Final Answer": "CORRECT/INCORRECT", 
"Confidence Score": <1-10>, 
"Winner": "<Agent Name>", 
"Reasoning": "<Explanation of decision>" }
Do not include any additional commentary.
\end{lstlisting}

\end{document}